\documentclass[conference]{IEEEtran}
\usepackage{times}
\usepackage{epsfig}
\usepackage{graphicx}
\usepackage{amsmath}
\usepackage{amssymb}

\ifCLASSINFOpdf
\else
\fi
\begin{document}
%
\title{Expression Recognition Using the Periocular Region: A Feasibility Study}

\author{\IEEEauthorblockN{Fernando Alonso-Fernandez}
\IEEEauthorblockA{\textit{School of ITE, Halmstad University} \\
Sweden \\
feralo@hh.se}
\and
\IEEEauthorblockN{Josef Bigun}
\IEEEauthorblockA{\textit{School of ITE, Halmstad University} \\
Sweden \\
josef.bigun@hh.se}
\and
\IEEEauthorblockN{Cristofer Englund}
\IEEEauthorblockA{\textit{RISE Viktoria, Gothenburg} \\
Sweden \\
cristofer.englund@hh.se}
}


%


\maketitle

\begin{abstract}
This paper investigates the feasibility of using the periocular region for expression recognition.
Most works have tried to solve this by analyzing  the whole face.
Periocular is the facial region in the immediate vicinity of the eye.
It has the advantage of being available over a wide range of distances and under partial face occlusion,
thus making it suitable for unconstrained or uncooperative scenarios.
We evaluate
five different image descriptors on a dataset of 1,574 images from 118 subjects.
The experimental results show an average/overall accuracy
of 67.0/78.0\% by fusion of several descriptors.
While this accuracy is still behind that attained with full-face methods,
it is noteworthy to mention that our initial approach employs only one frame to predict
the expression, in contraposition to
state of the art,
exploiting several order more data comprising spatial-temporal data which is often not available.
\end{abstract}


\begin{IEEEkeywords}
Expression Recognition, Emotion Recognition, Periocular Analysis, Periocular Descriptor.
\end{IEEEkeywords}

%
\IEEEpeerreviewmaketitle

\section{Introduction}

The ubiquitous computing paradigm is becoming a reality,
with an automation level in which people and devices interact seamlessly.
Ironically, one of the main challenges is the difficulty of users interacting
with these systems due to their increasing complexity \cite{[Pantic08]}.
Endowing machines with the ability to be aware of user emotions
(especially frustration, fear or dislike)
is thus of major importance for the next generation of user interfaces.

%
While it can be relatively easy for humans to recognize emotions and expressions,
achieved even unconsciously, extending such capability to machines is a very challenging
task that is attracting great interest.
Visible manifestations (face, gestures, etc.) \cite{[Soleymani16]}
are popular since they use
the same cues that humans rely upon,
with most human beings displaying similar
manifestations in response to identical emotional stimuli.
They also have the advantage that can be captured relatively easy with cameras,
even without active cooperation.
Other research make use of bio-signals such as
electroencephalography (EEG) \cite{[Menezes17]}
or electrocardiography (ECG) \cite{[Agrafioti12]},
but these require higher cooperation
due to the use of electrodes attached to the body or head,
and are not the focus of this paper.

Facial emotions are usually categorised into six classes:
anger, sad, surprise,
happy, disgust, fear \cite{[Ekman71]}.
Some recent works consider a most extensive list
of e.g. up to 26 categories \cite{[Kosti17]}.
%
%
A number of approaches have been proposed during the
last decade, e.g. \cite{[Lucey10]},
\cite{[Sariyanidi15]},
\cite{[Zhao16]},
\cite{[Desrosiers16]},
which can be broadly divided into
geometric
shape-based methods and appearance-based methods.
Geometric
shape-based methods use facial landmark information
alone, since expressions are defined by their relative position,
but performance of these methods is dependant of a
reliable extraction of facial landmarks.
Appearance based-methods
employ
texture information such as Local Binary Patterns (LBP) \cite{[Ojala02]}
or Scale Invariant Feature Transform (SIFT) features \cite{[Lowe04]}.
More progress has been achieved recently due to the
emergence of deep learning methods \cite{[Lopes17]}.
%
%

\begin{figure}[t]
\centering
\includegraphics[width=0.46\textwidth]{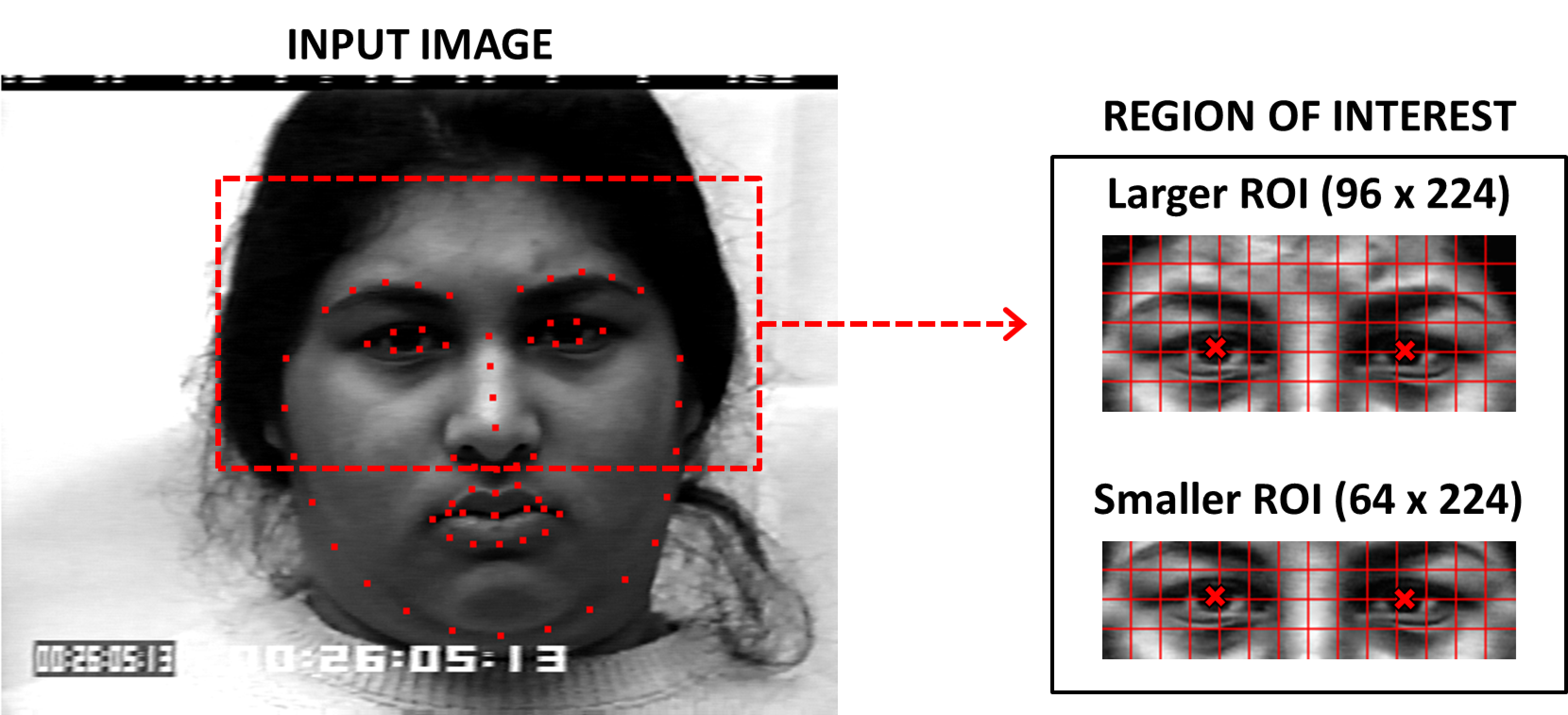}
\caption{Extraction of the region of interest. Blocks shown are of size 16$\times$16 pixels.}
\label{fig:ROI}
\end{figure}

Most of the previous research
in recognizing emotions are focused on face analysis.
A few papers use other visual clues such as
the location of shoulders \cite{[Nicolaou11]},
the body pose \cite{[Schindler08a]},
or the visual context \cite{[Kosti17]}.
Here, we tackle this issue by analyzing images from the periocular region.
Periocular refers to the facial region in the vicinity of the eye,
including eyelids, lashes and eyebrows (see Figure~\ref{fig:ROI}).
To the
best of our knowledge, this is the first study using periocular images
for the task of predicting expressions.
This region can be acquired largely relaxing the acquisition conditions,
in contraposition to the more carefully controlled conditions
usually needed in face,
making it suitable for unconstrained and uncooperative scenarios.
The periocular region has been shown as one of the most
discriminative regions of the face, with previous research
showing its impressive capabilities for tasks such as
person identity, gender or ethnicity classification \cite{[Alonso16]}.




\begin{table}[htb]
\small
\begin{center}
\begin{tabular}{ccccc}

 \multicolumn{2}{c}{} \\

\multicolumn{1}{c}{\textbf{system}} &
 \multicolumn{1}{c}{\textbf{size}} & & \multicolumn{1}{c}{\textbf{expression}} &
 \multicolumn{1}{c}{\textbf{number}} \\ \cline{1-2} \cline{4-5}

\multicolumn{1}{c}{LBP} & 8 &  & neutral & 593  \\ \cline{1-2} \cline{4-5}

\multicolumn{1}{c}{HOG} & 8 &  & angry &  135 \\ \cline{1-2} \cline{4-5}

\multicolumn{1}{c}{GABOR} & 30 &  & contempt &  54  \\ \cline{1-2} \cline{4-5}

\multicolumn{1}{c}{GLCM} & 5 &  & disgust & 177  \\ \cline{1-2} \cline{4-5}

\multicolumn{1}{c}{GIST} & 32 &  & fear &  75  \\ \cline{1-2} \cline{4-5}

 &  &  & happy &  207  \\ \cline{4-5}

 &  &  & sad &  84  \\ \cline{4-5}

 &  &  & surprise &  249  \\ \cline{4-5}

\multicolumn{2}{c}{} \\

\end{tabular}

\end{center}
\caption{Left: Size of the feature vectors per image block. Right: Number of available images per expression.}
\label{tab:vector-size}
\end{table}
\normalsize

\section{Expression Recognition}


This section describes the five feature extraction methods evaluated.
Features are extracted in the region of interest that is defined with respect to the eyes
(see Figure~\ref{fig:ROI}).
The periocular image is decomposed into non-overlapped blocks.
Features are extracted separately on each block, and then concatenated into a single vector, which
constitutes the image feature vector.
\\

\noindent \textbf{Local Binary Patterns (LBP)} \cite{[Ojala02]}.
For each pixel $p$, a $3 \times 3$ neighborhood is considered. Every
neighbor $p_i$ ($i$=1...8) is 
assigned a
binary value of 1 if $p_i>p$, or 0 otherwise. The binary values are
then concatenated into a 8-bits binary number, and the decimal
equivalent is assigned to characterize the texture at $p$, leading
to 2$^8$=256 possible labels.
The LBP values of all pixels within a given block are then
quantized into a 8-bin histogram,
and the histogram is further normalized to
account for local illumination and contrast variations.
%
\\

\textbf{Histogram of Oriented Gradients (HOG)} \cite{[Dalal05]}.
%
The gradient orientation and magnitude are computed in each
pixel. The histogram of orientations is then built, with each bin
accumulating corresponding gradient magnitudes.
The HOG values of all pixels within a given block are then
quantized into a 8-bin histogram,
and the histogram is further normalized to
account for local illumination and contrast variations. 
%
\\

\noindent \textbf{Gabor Features (GABOR)}.
Gabor filters are texture filters selective in frequency
and orientation.
Here, the local power
spectrum is sampled at the center of each block by a set of 30
Gabor filters organized in 5 frequency and 6 orientation channels
\cite{[Alonso15]}.
This sampling sparseness allows direct filtering in the
image domain without needing the Fourier transform, with significant
computational savings.
Filters are equally spaced in the log-polar frequency
plane, achieving full coverage of the spectrum.
The magnitude of Gabor responses of all frequency and orientation channels
are then grouped into a single complex vector.
\\

\noindent \textbf{Gray Level Co-occurrence Matrix (GLCM)}.
The GLCM is a joint probability distribution function of gray level pairs in a given image $I(p,q)$ \cite{[Haralick73]}.
Each element $C(i,j)$ in the GLCM specifies the probability that a pixel with intensity
value $i$ occurs in the image $I(p,q)$ at an offset $d=(\Delta p,\Delta q)$ of a pixel with intensity value $j$.
Usually the computation is done between neighboring pixels (i.e. $\Delta p=1$ or $\Delta q=1$).
To achieve rotational invariance, the GLCM is computed using a set of offsets uniformly covering the
0-180 degrees range (e.g. 0, 45, 90 and 135 degrees).
Once the GLCM is computed, various texture features are extracted and averaged across the different orientations.
The original paper \cite{[Haralick73]} defined 14 measures of textural features,
however they show certain redundancies.
For this reason, we employ a representative selection,
including the following features: contrast, homogeneity, entropy, energy, and autocorrelation.
\\

\textbf{Perceptual descriptors (GIST)} \cite{[Oliva01]}.
This consists of five perceptual dimensions related with scene
description, correlated with the second-order statistics and spatial
arrangement of structured image components:
\emph{naturalness}, which quantizes the vertical and horizontal edge
distribution; \emph{openness}, presence or lack of reference points;
\emph{roughness}, size of the largest prominent object;
\emph{expansion}, depth of the space gradient; and
\emph{ruggedness}, which quantizes the contour orientation that
deviates from the horizontal.
%
%
These descriptors are extracted by applying a set of
32 Gabor filters organized in 4 frequency and 8 orientation channels.
Each Gabor filter is applied to the entire block, and the magnitude of
all pixel responses are then averaged.
%
%
Prior to feature extraction, image blocks are pre-filtered via local
normalization of the intensity variance, in order to reduce
illumination effects and prevent some local regions to dominate the
energy spectrum.

\begin{table*}[htb]
	\begin{center}
			\begin{tabular}[htb]{|c|c|c|c|c||c|c|c|c|c|c|c||c|c|c|}
				
				\multicolumn{1}{c}{}\\
				
				\multicolumn{2}{c}{} & \multicolumn{6}{c}{Large periocular region} & \multicolumn{1}{c}{} &\multicolumn{6}{c}{Small periocular region} \\ \cline{3-8} \cline{10-15}

                \multicolumn{2}{c}{} & \multicolumn{3}{|c||}{16$\times$16 blocks} & \multicolumn{3}{c|}{32$\times$32 blocks} & \multicolumn{1}{|c|}{} & \multicolumn{3}{c||}{16$\times$16 blocks} & \multicolumn{3}{c|}{32$\times$32 blocks}\\  \cline{3-8} \cline{10-15}
				
				\multicolumn{1}{c}{} & \multicolumn{1}{c|}{} & Average & Overall & Min & Average  & Overall & Min
                & \multicolumn{1}{c|}{} & Average  & Overall & Min  & Average  & Overall & Min  \\

        		\multicolumn{1}{c}{Feature} & \multicolumn{1}{c|}{} &  Acc. &  Acc. &  Acc. &  Acc. &  Acc. &  Acc.
                & \multicolumn{1}{c|}{} &  Acc. &  Acc. &  Acc. &   Acc. &  Acc. & Acc. \\ \cline{1-1} \cline{3-8} \cline{10-15}

                LBP & & 59.8\% & 69.6\% & 28.0\% & 53.6\% & 59.7\% & 29.3\% &  & 59.0\% & 68.9\% & 21.3\% &  53.0\% & 59.0\%  & 28.0\% \\ \cline{1-1}  \cline{3-8} \cline{10-15}

                HOG & & 62.7\% & 72.0\% & 33.3\% & 55.6\% & 63.8\% & 29.3\% &  & 61.0\% & 68.7\% & \textbf{34.7\%} &  53.7\% & 62.4\%  & 18.7\% \\ \cline{1-1}  \cline{3-8} \cline{10-15}

                GABOR & & 62.5\% & \textbf{75.5\%} & 16.7\% &  \textbf{60.5\%} & \textbf{70.7\%} & \textbf{31.0\%} &  & 60.2\% & 72.9\% & 16.7\% &  \textbf{58.0\%} & \textbf{66.7\%} & 30.7\%  \\ \cline{1-1}  \cline{3-8} \cline{10-15}
				
                GLCM & & 56.1\% & 64.2\% & 22.7\% & 48.0\% & 55.0\% & 14.7\% &  & 52.2\% & 61.8\% & 20.2\% &  45.6\% & 54.0\% & 16.0\%  \\ \cline{1-1}  \cline{3-8} \cline{10-15}
				
                GIST & & \textbf{63.5\%} & 73.8\% & \textbf{34.5\%}  & 57.4\% & 65.4\% & 28.0\%  & & \textbf{63.4\%} & \textbf{73.3\%} & 27.8\%  & \textbf{58.1\%} & \textbf{66.7\%} & \textbf{33.3\%}  \\ \cline{1-1}  \cline{3-8} \cline{10-15}

                \multicolumn{1}{c}{}\\

			\end{tabular}
		\caption{Classification accuracy of the different features employed. \label{tab:accuracy-all}}
	\end{center}
\end{table*}
\normalsize

\section{Experimental Results}

\subsection{Database}

\begin{figure}[t]
\centering
\includegraphics[width=0.46\textwidth]{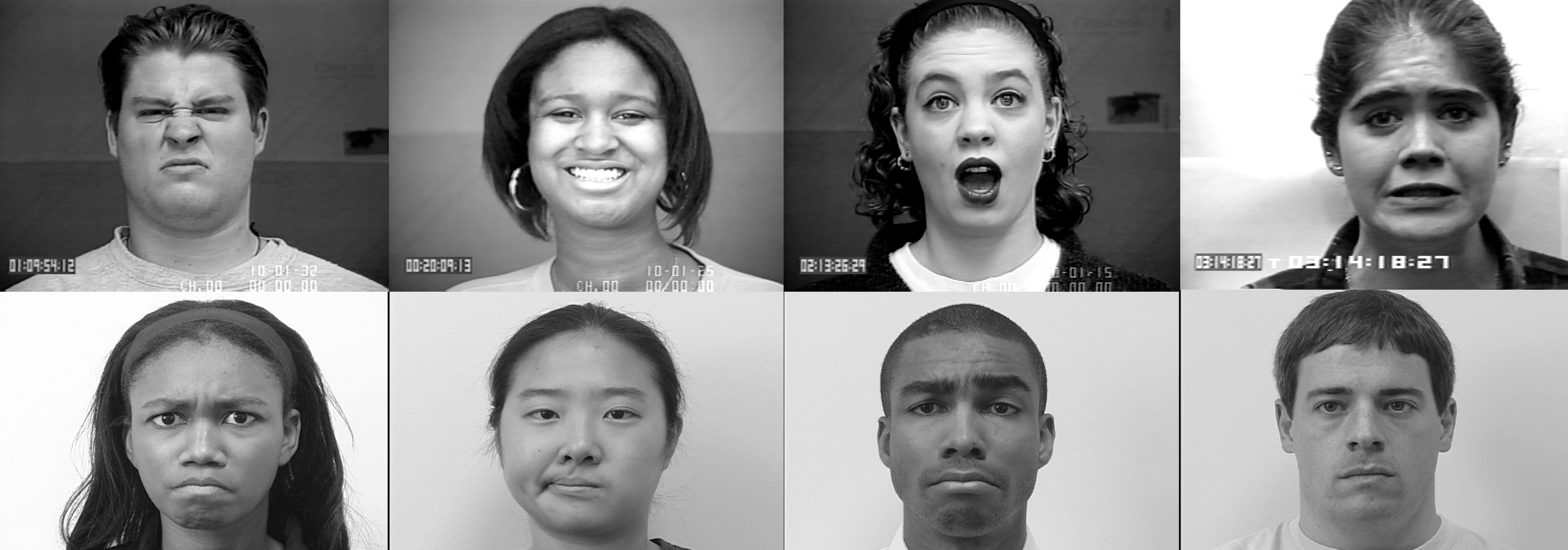}
\includegraphics[width=0.46\textwidth]{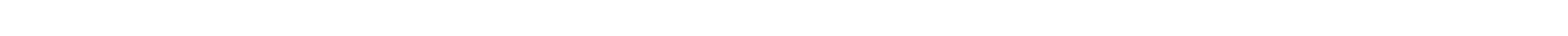}
\includegraphics[width=0.46\textwidth]{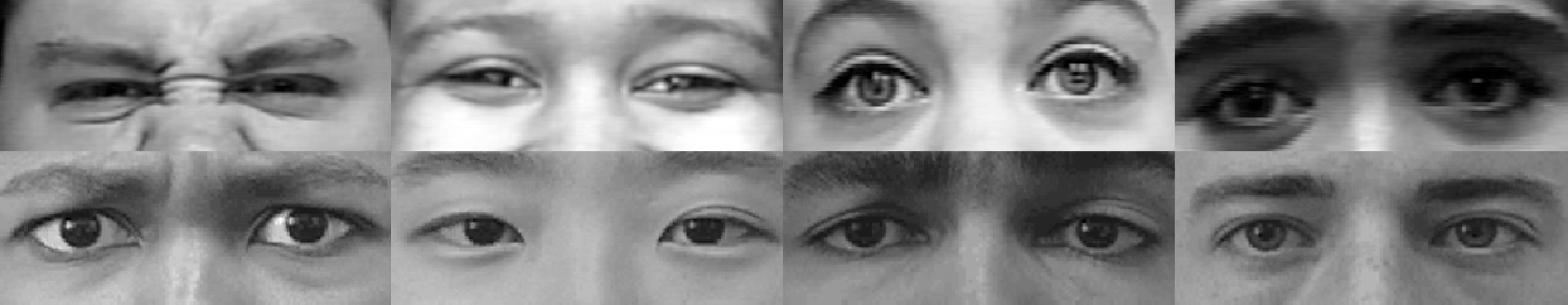}
\caption{Examples from CK+ database \cite{[Lucey10]} (from left to right and
top to bottom): disgust, happy, surprise, fear, angry, contempt, sadness, and neutral.
Top image: whole face. Bottom image: corresponding periocular region.}
\label{fig:CKdb}
\end{figure}

We use the Extended Cohn-Kanade Dataset (CK+) 
\cite{[Lucey10]}
to evaluate the proposed features.
This database (see Figure~\ref{fig:CKdb})
is widely used to benchmark emotion recognition methods.
It contains 123 different subjects and 593 frontal videos. Among these, 327
videos (corresponding to 118 subjects) are labelled with the seven basic emotion categories
(disgust, happy, surprise, fear, angry, contempt, and sadness).
Duration varies from 7 to 60 frames,
with all sequences beginning at the onset (neutral emotion) in the first frame,
and stoping at the apex (peak expression emotion) in the last frame.
The sequences are annotated with 68 landmark points of the face
(Figure~\ref{fig:ROI}, left).
The key-frames of each video sequence are manually labelled, and the remaining
frames are estimated by Active Appearance Models (AAM) \cite{[Cootes01]}.

\subsection{Protocol}

We resize each frame via bicubic interpolation to have a constant eye-to-eye
distance of 100 pixels (average value of the database).
The center of each eye is computed as the centroid of the 6 available
landmarks (Figure~\ref{fig:ROI}, left).
Rotation is compensated based on the straight line that crosses the two eye
centers.
Then, a rectangular region of interest is extracted around the eyes.
Images are further equalized with CLAHE \cite{[Zuiderveld94clahe]} to compensate
local illumination variability.
We consider two sizes of the region of interest:
$i$) a small region of 64$\times$224, and
$ii$) a large region of 96$\times$224.
In the first case, the eyebrows might not be included, while the second case is
chosen to ensure that the eyebrows and part of the forehead is included
(Figure~\ref{fig:ROI}, right).
In both cases, we test blocks of size 16$\times$16 and 32$\times$32 pixels.
Size of extracted feature vectors (per block) is given in Table~\ref{tab:vector-size}, left.

We select the first image of each video as neutral,
and the last three frames as the expression indicated in the label.
The number of available images per expression can be seen in Table~\ref{tab:vector-size}, right.
Prediction is done separately on each image, so each video contributes
with four different images to be classified.
We use linear Support Vector Machines (SVMs)
to perform the expression classification \cite{[Vapnik95]}.
A one-vs-one multi-class approach is used. For every feature and $N$ classes,
$N(N-1)/2$ binary SVM classifiers are used. Classification is made based on which class has most
number of binary classifications made towards it (voting scheme).
Following the work in \cite{[Lucey10]}, we employ the leave-one-subject-out
protocol in our experiments, resulting in 118 different training and test sets.
Training images are mirrored in the horizontal direction to duplicate the size
of the training set.

\subsection{Classification Accuracy}

The overall and average accuracies of each feature extraction method
are listed in Table~\ref{tab:accuracy-all}.
The average accuracy is computed by the mean of each category accuracy.
On the other hand, the overall accuracy is determined by the percentage
of images that are correctly classified over all tests.
We also report the minimum accuracy, which refers to the accuracy
of the worst performing category.
From Table~\ref{tab:accuracy-all}, we can observe that:
\begin{itemize}
  \item If we look at the average or overall accuracies,
    the impact of using a small periocular region is negligible in most cases,
    meaning that the forehead region does not play an important role in classifying expressions.
    Reductions in accuracy due to a smaller region are of 4-5\% at most.
    The minimum accuracy, on the other hand, is more affected in some cases
    (e.g. LBP: 28.0\% to 21.3\% with 16$\times$16 blocks;
        HOG: 29.3\% to 18.7\% with 32$\times$32 blocks; or
        GIST: 34.5\% to 27.8\% with 16$\times$16 blocks).
\end{itemize}

\begin{itemize}
  \item In general, using 16$\times$16 blocks results in better accuracy than
  using 32$\times$32 blocks.
  It is expected that displacements of the periocular skin and landmarks
  (eyelids, lashes, brows, etc.) are better captured with smaller blocks.
  This is because skin and landmark movements will transfer sooner
  to adjacent blocks if their size is sufficiently small.
  Using smaller blocks on the one hand results in a bigger feature vector.
\end{itemize}

\begin{itemize}
  \item The best performing descriptors are GABOR and GIST.
  Both are based on Gabor filters,
  and have the biggest vector size (Table~\ref{tab:vector-size}).
  The GABOR descriptor only applies Gabor filters to the center of each block,
  while the GIST descriptor makes the convolution with the entire block.
  Despite this different sparseness, their performance is very similar,
  meaning that it is sufficient to sample only one point per image block.
  It is also noteworthy to mention that the other three descriptors have a vector whose
  size is one order of magnitude smaller, but their performance in some cases is not far away
  from GABOR or GIST (e.g. HOG has an accuracy similar to GABOR in some cases).
  The worst performing method is GLCM, but it only employs 5 features per block.
\end{itemize}

Based on the results of Table~\ref{tab:accuracy-all}, we select the case of
large periocular region and blocks of size 16$\times$16
for further consideration, since this is overall the best performing combination.
The results for each expression category are shown in Table~\ref{tab:accuracy-classes}.
It can be observed that all methods achieves better performance for
high energy expressions such as `happy', `disgust' and `surprise'.
This is also observed in other studies \cite{[Zhao16],[Lucey10]}.
These are the expressions exhibiting large displacements of the facial landmarks.
In these cases, GABOR and GIST descriptors achieve accuracies higher than 80\%.
The neutral expression is also among the best performing one.
Other expressions such as `fear' and `sad', which are referred as more complex,
usually shows worse performance \cite{[Zhao16]}. In our experiments, the accuracy
of these is not above 45\% for any given feature.
It should be noted however that these classes are the most misrepresented
in the CK+ database (Table~\ref{tab:vector-size}, right).

We then perform fusion experiments of some of the available features.
Results are given
in Tables~\ref{tab:accuracy-fusion-all} and \ref{tab:accuracy-fusion-classes}.
Since GABOR and GIST are the best performing features, we
test the fusion of GABOR+all other features, and GIST+all other features,
The complementarity of these two best methods is also tested with the fusion
GABOR+GIST.
We also test the combination of lightweight methods: LBP+HOG+GLCM (i.e. those with the
smallest feature vector, see Table~\ref{tab:vector-size}).
It is observed that feature fusion produces additional improvements w.r.t. the
individual features alone. For example, the best average accuracy is brought from 63.5\%
(Table~\ref{tab:accuracy-all}) to 67.0\%, while the overall accuracy goes
from 75.5\% to 78.0\%.
It should be remarked however that for example, the lightweight combination of LBP+HOG+GLCM
has a performance which is only slightly worse.
From Table \ref{tab:accuracy-fusion-classes},
we can also observe that expressions such as `angry' are brought from an accuracy of 65.9\%
(Table~\ref{tab:accuracy-classes}) to 71.9\%, although `fear', `sad' or `contempt' does not improve with the fusion.

\begin{table}[htb]
	\scriptsize
	\begin{center}
			\begin{tabular}[htb]{|c|c|c|c|c|c|c|c|c|}
				
				\multicolumn{1}{c}{}\\ \cline{2-9}
				
				\multicolumn{1}{c|}{\textbf{LBP}} & Neu & Ang & Con & Dis & Fea & Hap & Sad & Sur \\ \hline

Neutral & \textbf{74.3} & 5.2 & 4.3 & 1.5 & 0.9 & 6.7 & 3.1 & 4 \\ \hline
Angry & 8.9 & \textbf{61.5} & 1.5 & 8.9 & 3 & 12.6 & 3.7 & 0 \\ \hline
Contempt & 53.7 & 0 & \textbf{40.7} & 0 & 0 & 0 & 1.9 & 3.7 \\ \hline
Disgust & 5.1 & 4.5 & 0 & \textbf{77.4} & 0 & 13 & 0 & 0 \\ \hline
Fear & 9.3 & 6.7 & 4 & 2.7 & \textbf{28} & 6.7 & 17.3 & 25.3 \\ \hline
Happy & 10.1 & 5.3 & 0 & 4.3 & 0.5 & \textbf{77.8} & 1.4 & 0.5 \\ \hline
Sad & 26.2 & 14.3 & 4.8 & 0 & 10.7 & 4.8 & \textbf{32.1} & 7.1 \\ \hline
Surprise & 4.4 & 0 & 0.8 & 0 & 7.2 & 0 & 0.8 & \textbf{86.7} \\ \hline

                \multicolumn{1}{c}{}\\

                \multicolumn{1}{c}{}\\ \cline{2-9}
				
				\multicolumn{1}{c|}{\textbf{HOG}} & Neu & Ang & Con & Dis & Fea & Hap & Sad & Sur \\ \hline

Neutral & \textbf{77.7} & 4.3 & 4.3 & 1.2 & 1.2 & 4.9 & 4.3 & 2.1 \\ \hline
Angry & 14.1 & \textbf{55.6} & 0 & 16.3 & 3.7 & 8.1 & 2.2 & 0 \\ \hline
Contempt & 48.1 & 1.9 & \textbf{33.3} & 0 & 0 & 3.7 & 9.3 & 3.7 \\ \hline
Disgust & 5.6 & 7.3 & 0.6 & \textbf{78} & 0 & 5.6 & 2.8 & 0 \\ \hline
Fear & 6.7 & 4 & 1.3 & 2.7 & \textbf{44} & 1.3 & 17.3 & 22.7 \\ \hline
Happy & 10.1 & 4.8 & 0 & 6.3 & 0 & \textbf{78.7} & 0 & 0 \\ \hline
Sad & 20.2 & 14.3 & 7.1 & 0 & 11.9 & 0 & \textbf{44} & 2.4 \\ \hline
Surprise & 5.2 & 0 & 1.6 & 0 & 3.2 & 0 & 0 & \textbf{90} \\ \hline

                \multicolumn{1}{c}{}\\

                \multicolumn{1}{c}{}\\ \cline{2-9}
				
				\multicolumn{1}{c|}{\textbf{GABOR}} & Neu & Ang & Con & Dis & Fea & Hap & Sad & Sur \\ \hline

Neutral & \textbf{88.7} & 1.5 & 0.9 & 1.2 & 1.2 & 2.8 & 1.8 & 1.8 \\ \hline
Angry & 8.9 & \textbf{65.9} & 0.7 & 10.4 & 7.4 & 5.2 & 1.5 & 0 \\ \hline
Contempt & 68.5 & 0 & \textbf{16.7} & 0 & 0 & 1.9 & 1.9 & 11.1 \\ \hline
Disgust & 7.3 & 5.6 & 0 & \textbf{82.5} & 1.7 & 2.8 & 0 & 0 \\ \hline
Fear & 17.3 & 12 & 2.7 & 5.3 & \textbf{29.3} & 4 & 5.3 & 24 \\ \hline
Happy & 15.9 & 0 & 0 & 1.9 & 1.4 & \textbf{79.7} & 0 & 1 \\ \hline
Sad & 31 & 7.1 & 3.6 & 1.2 & 8.3 & 0 & \textbf{45.2} & 3.6 \\ \hline
Surprise & 6.4 & 0 & 1.6 & 0 & 0.4 & 0 & 0 & \textbf{91.6} \\ \hline

                \multicolumn{1}{c}{}\\

                \multicolumn{1}{c}{}\\ \cline{2-9}
				
				\multicolumn{1}{c|}{\textbf{GLCM}} & Neu & Ang & Con & Dis & Fea & Hap & Sad & Sur \\ \hline

Neutral & \textbf{65.4} & 4.9 & 4.6 & 1.5 & 6.1 & 5.5 & 4.9 & 7 \\ \hline
Angry & 8.1 & \textbf{57} & 0 & 19.3 & 3.7 & 5.9 & 4.4 & 1.5 \\ \hline
Contempt & 25.9 & 7.4 & \textbf{46.3} & 0 & 0 & 0 & 1.9 & 18.5 \\ \hline
Disgust & 6.2 & 9 & 0 & \textbf{78.5} & 2.8 & 3.4 & 0 & 0 \\ \hline
Fear & 16 & 9.3 & 0 & 12 & \textbf{22.7} & 4 & 9.3 & 26.7 \\ \hline
Happy & 9.2 & 4.8 & 0.5 & 3.4 & 5.8 & \textbf{71} & 1.4 & 3.9 \\ \hline
Sad & 33.3 & 9.5 & 2.4 & 2.4 & 15.5 & 1.2 & \textbf{28.6} & 7.1 \\ \hline
Surprise & 10.4 & 0 & 0 & 0.4 & 5.6 & 1.6 & 2.8 & \textbf{79.1} \\ \hline

                \multicolumn{1}{c}{}\\

                \multicolumn{1}{c}{}\\ \cline{2-9}
				
				\multicolumn{1}{c|}{\textbf{GIST}} & Neu & Ang & Con & Dis & Fea & Hap & Sad & Sur \\ \hline

Neutral & \textbf{82} & 2.8 & 3.1 & 1.2 & 2.1 & 2.8 & 4 & 2.1 \\ \hline
Angry & 8.9 & \textbf{64.4} & 4.4 & 11.9 & 2.2 & 4.4 & 3.7 & 0 \\ \hline
Contempt & 48.1 & 1.9 & \textbf{40.7} & 0 & 0 & 0 & 3.7 & 5.6 \\ \hline
Disgust & 7.3 & 6.2 & 0 & \textbf{80.8} & 0 & 4 & 1.7 & 0 \\ \hline
Fear & 9.3 & 9.3 & 0 & 0 & \textbf{34.7} & 4 & 17.3 & 25.3 \\ \hline
Happy & 7.2 & 1.9 & 0 & 5.3 & 1.4 & \textbf{84.1} & 0 & 0 \\ \hline
Sad & 21.4 & 21.4 & 0 & 3.6 & 15.5 & 0 & \textbf{34.5} & 3.6 \\ \hline
Surprise & 7.2 & 0 & 0 & 0 & 4.8 & 0 & 1.2 & \textbf{86.7} \\ \hline

                \multicolumn{1}{c}{}\\
			\end{tabular}
		\caption{Classification accuracy of each class (large periocular region, 16$\times$16 blocks). \label{tab:accuracy-classes}}
	\end{center}
\end{table}
\normalsize

\begin{table}[htb]
	\begin{center}
			\begin{tabular}[htb]{|c|c|c|c|}

				\multicolumn{1}{c}{}\\ \hline

				& Average & Overall & Min \\

        		Method & Acc. &  Acc. &  Acc.  \\ \hline

                LBP+HOG+GABOR+GLCM & 65.0\% & \textbf{78.0\%} & 16.7\% \\ \hline

                LBP+HOG+GLCM+GIST & \textbf{67.0\%} & 77.6\% & \textbf{40.5\%}  \\ \hline

                GABOR+GIST & 65.5\% & 77.6\% & 30.7\%  \\ \hline

                LBP+HOG+GLCM & 66.6\% & 77.6\% & 35.2\%   \\ \hline

                \multicolumn{1}{c}{}\\

			\end{tabular}
		\caption{Classification accuracy of different fusion combinations (large periocular region, 16$\times$16 blocks). \label{tab:accuracy-fusion-all}}
	\end{center}
\end{table}
\normalsize

\begin{table}[htb]
	\scriptsize
	\begin{center}
			\begin{tabular}[htb]{|c|c|c|c|c|c|c|c|c|}
				
				\multicolumn{1}{c}{}\\

                \multicolumn{1}{c}{} & \multicolumn{8}{c}{\textbf{LBP+HOG+GABOR+GLCM}}\\ \cline{2-9}
				\multicolumn{1}{c|}{} & Neu & Ang & Con & Dis & Fea & Hap & Sad & Sur \\ \hline

Neutral & \textbf{89} & 2.8 & 2.8 & 0.9 & 0.3 & 2.8 & 0.9 & 0.6 \\ \hline
Angry & 7.4 & \textbf{71.9} & 0 & 8.9 & 2.2 & 5.2 & 4.4 & 0 \\ \hline
Contempt & 66.7 & 0 & \textbf{16.7} & 0 & 0 & 1.9 & 0 & 14.8 \\ \hline
Disgust & 6.8 & 3.4 & 0 & \textbf{84.7} & 1.7 & 3.4 & 0 & 0 \\ \hline
Fear & 9.3 & 8 & 4 & 5.3 & \textbf{30.7} & 4 & 10.7 & 28 \\ \hline
Happy & 9.7 & 0 & 0 & 1.4 & 0 & \textbf{88.4} & 0 & 0.5 \\ \hline
Sad & 26.2 & 16.7 & 3.6 & 0 & 2.4 & 0 & \textbf{47.6} & 3.6 \\ \hline
Surprise & 6 & 0 & 1.2 & 0 & 1.6 & 0 & 0 & \textbf{91.2} \\ \hline

                \multicolumn{1}{c}{}\\

                \multicolumn{1}{c}{} & \multicolumn{8}{c}{\textbf{LBP+HOG+GLCM+GIST}}\\ \cline{2-9}
				\multicolumn{1}{c|}{} & Neu & Ang & Con & Dis & Fea & Hap & Sad & Sur \\ \hline

Neutral & \textbf{87.8} & 4.3 & 2.4 & 0.3 & 0.6 & 3.1 & 0.6 & 0.9 \\ \hline
Angry & 5.9 & \textbf{66.7} & 3 & 13.3 & 0.7 & 6.7 & 3.7 & 0 \\ \hline
Contempt & 55.6 & 0 & \textbf{40.7} & 0 & 0 & 0 & 0 & 3.7 \\ \hline
Disgust & 5.1 & 6.8 & 0 & \textbf{79.1} & 1.1 & 7.9 & 0 & 0 \\ \hline
Fear & 9.3 & 8 & 0 & 0 & \textbf{41.3} & 4 & 13.3 & 24 \\ \hline
Happy & 9.7 & 0 & 0 & 1.9 & 1.4 & \textbf{87} & 0 & 0 \\ \hline
Sad & 22.6 & 11.9 & 1.2 & 3.6 & 17.9 & 0 & \textbf{40.5} & 2.4 \\ \hline
Surprise & 4 & 0 & 0.4 & 0 & 2.8 & 0 & 0 & \textbf{92.8} \\ \hline

                \multicolumn{1}{c}{}\\

                \multicolumn{1}{c}{} & \multicolumn{8}{c}{\textbf{GABOR+GIST}}\\ \cline{2-9}
				\multicolumn{1}{c|}{} & Neu & Ang & Con & Dis & Fea & Hap & Sad & Sur \\ \hline

Neutral & \textbf{88.1} & 2.8 & 2.4 & 0.3 & 0.6 & 2.1 & 1.5 & 2.1 \\ \hline
Angry & 10.4 & \textbf{60.7} & 0.7 & 18.5 & 1.5 & 5.2 & 3 & 0 \\ \hline
Contempt & 57.4 & 0 & \textbf{33.3} & 0 & 0 & 0 & 0 & 9.3 \\ \hline
Disgust & 6.8 & 2.8 & 0 & \textbf{84.2} & 0 & 4.5 & 1.7 & 0 \\ \hline
Fear & 10.7 & 10.7 & 0 & 2.7 & \textbf{30.7} & 4 & 17.3 & 24 \\ \hline
Happy & 6.3 & 0 & 0 & 0 & 0.5 & \textbf{92.3} & 0 & 1 \\ \hline
Sad & 23.8 & 16.7 & 0 & 2.4 & 10.7 & 0 & \textbf{42.9} & 3.6 \\ \hline
Surprise & 6 & 0 & 0.8 & 0 & 1.6 & 0 & 0 & \textbf{91.6} \\ \hline

                \multicolumn{1}{c}{}\\

                \multicolumn{1}{c}{} & \multicolumn{8}{c}{\textbf{LBP+HOG+GLCM}}\\ \cline{2-9}
				\multicolumn{1}{c|}{} & Neu & Ang & Con & Dis & Fea & Hap & Sad & Sur \\ \hline

Neutral & \textbf{86.9} & 2.4 & 4.6 & 0.3 & 0.3 & 2.4 & 2.4 & 0.6 \\ \hline
Angry & 5.9 & \textbf{71.9} & 1.5 & 11.1 & 2.2 & 4.4 & 3 & 0 \\ \hline
Contempt & 53.7 & 0 & \textbf{35.2} & 0 & 0 & 1.9 & 0 & 9.3 \\ \hline
Disgust & 7.9 & 5.6 & 0 & \textbf{80.8} & 0 & 5.6 & 0 & 0 \\ \hline
Fear & 12 & 4 & 4 & 1.3 & \textbf{37.3} & 4 & 12 & 25.3 \\ \hline
Happy & 8.7 & 1 & 0 & 1.4 & 0 & \textbf{87.4} & 0 & 1.4 \\ \hline
Sad & 20.2 & 11.9 & 4.8 & 0 & 14.3 & 2.4 & \textbf{41.7} & 4.8 \\ \hline
Surprise & 5.6 & 0 & 1.2 & 0 & 1.6 & 0 & 0 & \textbf{91.6} \\ \hline

                \multicolumn{1}{c}{}\\

                \multicolumn{1}{c}{}\\
			\end{tabular}
		\caption{Classification accuracy of each class for different fusion combinations (large periocular region, 16$\times$16 blocks). \label{tab:accuracy-fusion-classes}}
	\end{center}
\end{table}
\normalsize

We finally show in Table \ref{tab:accuracy-face}
the accuracies reported by other expression recognition
studies on the CK+ database using the full face.
The overall accuracy in these cases range from 88.3\% to 96.6\%,
while our best accuracy using periocular images is of 78.0\%.
It should be considered however that these methods
carry out dynamic analysis, meaning that
expression is assessed by analyzing spatial-temporal changes
from the first (neutral) frame to the last (apex) frame of each video.
Therefore, they employ all the frames available in each video.
%
For this reason, they do not report results on the `neutral' expression,
since it is used as reference to measure the other seven expressions.
In this paper,
on the other hand, we focus on detecting the neutral and the apex
expression on individual frames.

\begin{table}[htb]
	\begin{center}
			\begin{tabular}[htb]{|c|c|c|c|}

				\multicolumn{1}{c}{}\\ \hline

				& Average & Overall & Min \\

        		Method & Acc. &  Acc. &  Acc.  \\ \hline

                Lucey et al. \cite{[Lucey10]} & 83.3\% & 88.3\% & 65.2\%  \\ \hline

                Chew et al. \cite{[Chew12]} & 89.4\% &  n/a & n/a \\ \hline

                Wang et al. \cite{[Wang13]} & 86.3\% & 88.8\% & 76.0\%   \\ \hline

                Liu et al. \cite{[Liu14a]} & 87.9\% & 92.4\% & 66.7\% \\ \hline

                Liu et al. \cite{[Liu16]} & 94.8\% & 96.6\% & 88.0\%  \\ \hline

                Zhao et al. \cite{[Zhao16]} & 93.5\% & 95.7\% & 82.1\%  \\ \hline

                Liu et al. \cite{[Liu17]} & 96.1\%  & n/a &  90.3\% \\ \hline

                \multicolumn{1}{c}{} \\

			\end{tabular}
		\caption{Recent expression recognition methods using the full face on the CK+ database. \label{tab:accuracy-face}}
	\end{center}
\end{table}
\normalsize

\section{Conclusion and Future Work}

In this paper, we evaluate the feasibility of using images of the periocular
region for expression recognition.
We employ five different descriptors for feature extraction,
and prediction is made with SVM classifiers.
Despite some studies have analyzed the impact of expression changes on the
performance of periocular recognition systems \cite{[Barroso16],[AlGashaam15]},
this is, to the best of our knowledge, the first study using periocular
images for the task of predicting expressions.
In this initial study, we carry out prediction on individual frames.
Based on the evaluation of a total of 1,574 still images from videos of
118 different subjects,
our results indicates the feasibility of using the periocular region as a
predictor of facial expression.
By fusion of several descriptors, we attain an average/overall accuracy
of 67.0/78.0\%.
While previous studies making use of the full face report average accuracies
in the range of 83-96\%, they make use of dynamic information of the face,
i.e. they employ all the
frames that goes from the onset (neutral expression) in the first frame,
to the apex (peak expression) in the last frame of each video.

Future work will involve incorporating the mouth region to our study.
It is also expected that exploiting spatial-temporal information
(the dynamics of periocular movements across several
frames) will produce considerable improvements in the performance.
This is because facial emotions are mainly generated by the
movements of the facial muscles.
We would also like to include Convolutional Neural Networks
to the study of periocular expression prediction \cite{[Liu17]}.


\section*{Acknowledgment}

Author F. A.-F. thanks the Swedish Research Council for funding his research. Authors acknowledge
the CAISR program and the SIDUS-AIR project of the Swedish Knowledge Foundation.



%
%
%

\bibliographystyle{IEEEtran}
\bibliography{fernando1}

\end{document}